\def\year{2021}\relax
\documentclass[letterpaper]{article} 
\usepackage{aaai21}  
\usepackage{times}  
\usepackage{helvet} 
\usepackage{courier}  
\usepackage[hyphens]{url}  
\usepackage{graphicx} 
\def\UrlFont{\rm}  
\usepackage{natbib}  
\usepackage{caption} 
\usepackage{adjustbox}
\usepackage{booktabs}
\usepackage{amssymb}
\usepackage{array}\newcolumntype{P}[1]{>{\centering\arraybackslash}p{#1}}
\usepackage{makecell}
\usepackage{multirow}
\usepackage{xcolor}

\title{Are We There Yet? Learning to Localize in Embodied Instruction Following}
\author {
    Shane Storks\textsuperscript{\rm 1}\thanks{Work performed during an internship with Amazon Alexa AI.},
    Qiaozi Gao\textsuperscript{\rm 2},
    Govind Thattai\textsuperscript{\rm 2},
    Gokhan Tur\textsuperscript{\rm 2}\\
}
\affiliations {
    \textsuperscript{\rm 1} University of Michigan \ \ \textsuperscript{\rm 2} Amazon Alexa AI \\
    sstorks@umich.edu, \{qzgao, thattg, gokhatur\}@amazon.com
}

\begin{document}

\maketitle

\begin{abstract}
Embodied instruction following is a challenging problem requiring an agent to infer a sequence of primitive actions to achieve a goal environment state from complex language and visual inputs. Action Learning From Realistic Environments and Directives (ALFRED) is a recently proposed benchmark for this problem consisting of step-by-step natural language instructions to achieve subgoals which compose to an ultimate high-level goal. Key challenges for this task include localizing target locations and navigating to them through visual inputs, and grounding language instructions to visual appearance of objects. To address these challenges, in this study, we augment the agent’s field of view during navigation subgoals with multiple viewing angles, and train the agent to predict its relative spatial relation to the target location at each timestep. We also improve language grounding by introducing a pre-trained object detection module to the model pipeline. Empirical studies show that our approach exceeds the baseline model performance.
\end{abstract}

\section{Introduction}
As in-home robots become a reality, there are still fundamental problems along the intersection of vision, language, and robotics to be solved. A particularly challenging problem is embodied instruction following, where an agent must complete a task by following a human teacher's language instructions. The language instructions may require the agent to navigate through a space, and manipulate objects in the space. Thus, navigation skills and language grounding are essential capabilities for such agents, and this research area has received significant attention in recent years.

While much progress has been made toward this in graph-structured, discrete navigation environments \cite{mattersim,Zhu_2020_CVPR,majumdarImprovingVisionandLanguageNavigation2020}, the problem remains challenging in continuous navigation environments.
Action Learning From Realistic Environments and Directives (ALFRED) is a recently introduced benchmark dataset for embodied task learning, providing human-written natural language instructions for an agent to perform tasks in a virtual environment with continuous navigation \cite{shridharALFREDBenchmarkInterpreting2020}. We focus our work on this challenging benchmark.

\begin{figure}
    \centering
    \includegraphics[width=0.46\textwidth]{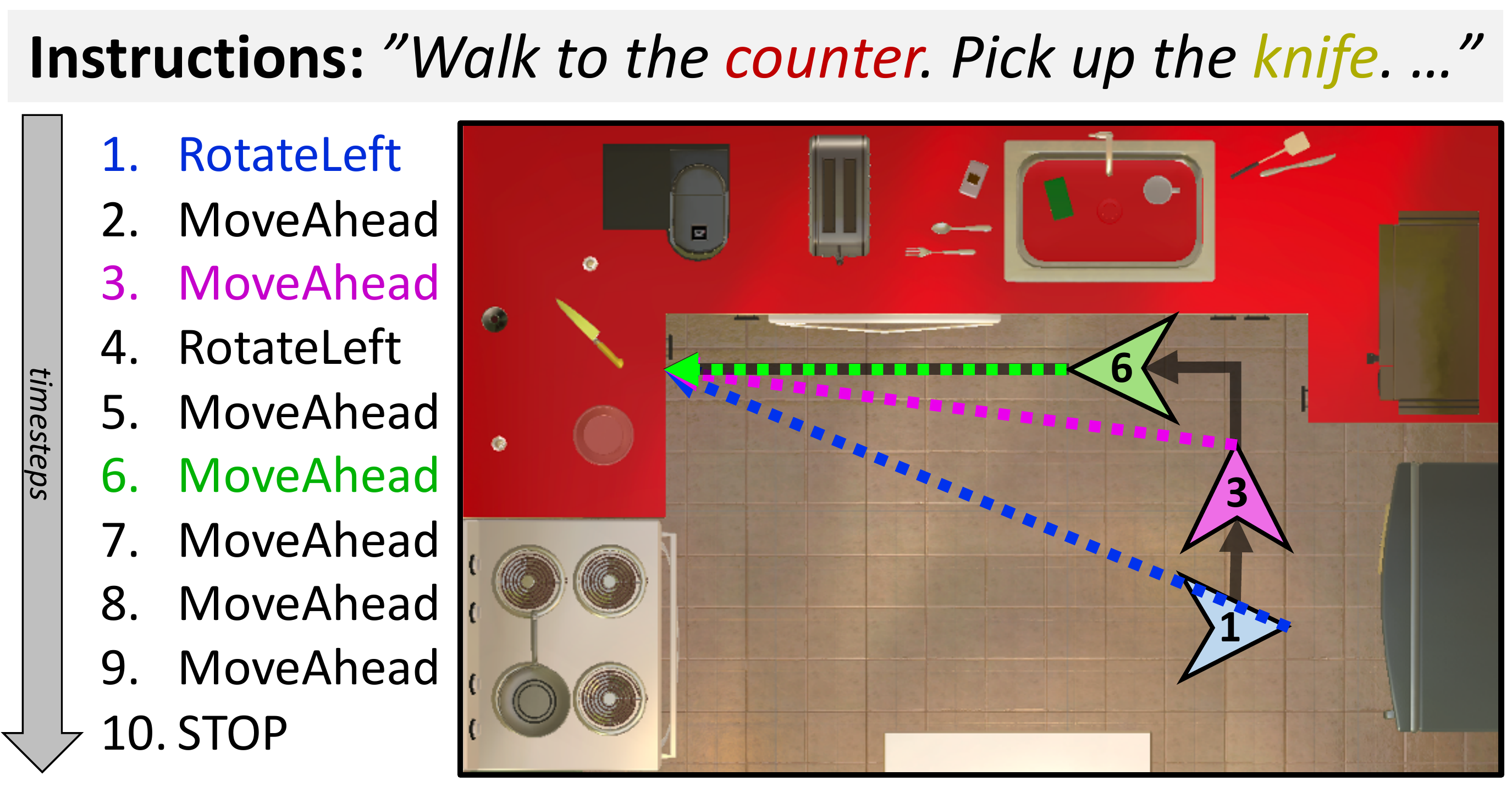}
    \caption{To improve an embodied agent's navigation capability, we use enhanced inputs from the environment to track the direction of the goal location from the agent's perspective at every timestep (colored dotted lines). This signal guides the agent during navigation. }
    \label{fig:map}
\end{figure}

To address key weaknesses of existing approaches for ALFRED, we implement several enhancements in this work. First, we train the model on a step-by-step granularity of language instructions. Next, we augment the inputs for navigation by adding panoramic visual observations at each timestep of navigation, and object segmentation masks of all objects in the environment. We use these inputs to train an object detection module that can be used at inference time. Lastly, we propose a novel, transformer-based localization module which we train to predict the agent's orientation angle with respect to the goal location. This module is then used at inference time to guide the agent during navigation, as shown in Figure~\ref{fig:map}. From these modifications, we show some improvements over a strong baseline approach.

\section{Related Work}

\paragraph{Language, vision, and robotics.}
Within the intersection of language, vision, and robotics, there are several threads of work toward embodied instruction following, navigation, and localization within a virtual environment. As such problems rely on the availability of virtual environments and large-scale data, many of these threads have spawned from the release of benchmarks and challenge datasets. \citet{changMatterport3DLearningRGBD2017}, \citet{ai2thor}, \citet{yan2018chalet}, \citet{habitat19iccv}, and \citet{xiagibson2019} propose various virtual environments for embodied AI research, each with their own considerations and features. \citet{embodiedqa} framed the traditional task of QA in a virtual environment, posing questions that required an agent to explore an environment to find the answer. \citet{reverie} proposes a remote object grounding challenge, where an embodied agent must find a particular object in a virtual environment. \citet{igibson} frames robotic motion planning in a virtual environment with movable objects as a reinforcement learning problem. \citet{mattersim} proposes the vision-and-language navigation (VLN) task, where an agent must use human-written language instructions to reach a goal location in a photorealistic virtual environment, and \citet{thomason:arxiv19} expands this task to a dialog-based formulation.

\paragraph{Vision-and language navigation.}
A particularly substantial body of work toward navigation in embodied instruction following exists in the VLN thread. \citet{friedSpeakerFollowerModelsVisionandLanguage2018} introduced a pragmatic reasoning module to the problem that learned to re-tell candidate navigation routes in natural language, and use this output to re-rank the routes. \citet{jain-etal-2019-stay} introduced a new graph-based metric for navigation fidelity, and used this as a new training objective to improve the fidelity of agents to navigation instructions. \citet{ma2019selfmonitoring} introduced another auxiliary training objective for the agent to estimate its progress toward the goal location, and \citet{ma2019theregretful} expanded upon this to give the agent the ability to backtrack during navigation. \citet{hu-etal-2019-looking} improve language grounding in navigation agents by using a pre-trained object detection system to augment visual inputs with localized representations of objects in the scene. \citet{Zhu_2020_CVPR} combined several previously proposed and novel self-supervised auxiliary training objectives for VLN to achieve state-of-the-art performance on the task. \citet{majumdarImprovingVisionandLanguageNavigation2020} pre-trained a transformer on a large-scale language grounding task to further improve performance.

\section{ALFRED Benchmark and Baseline}
In this work, we focus on the ALFRED benchmark. ALFRED is a recent embodied task learning benchmark where an agent must complete household tasks to achieve a specific goal state in a virtual environment powered by AI2 Thor~\cite{ai2thor}. First, we introduce the problem and the existing baseline approach formally.

\subsection{Problem Formulation}
An instance of ALFRED consists of a high-level \textit{goal} $G$ which consists of a sequence of $N$ \textit{subgoals} $g_i \in G$. Each subgoal may be a \textit{navigation} subgoal or a \textit{manipulation} subgoal, e.g., to pick up, put down, heat, or cool an object. Navigation subgoals consist of primitive navigation actions, e.g., to turn, move forward, or change the vertical heading angle, while manipulation subgoals consist mostly of primitive manipulation actions, e.g., to pick up, slice, or toggle an object. A typical ALFRED instance consists of alternating navigation and manipulation subgoals, and all primitive actions are performed sequentially.

These subgoals, when completed in order, achieve a goal state $S^*$ in the virtual environment. The internal representation of goals and subgoals are not provided for inference. Instead, the agent needs to reason from the associated human annotated natural language instructions. From language instructions $L_G$ and $L_{i}, i \in \{ 1, 2, \cdots, N \} $ describing the goal and subgoals respectively, as well as visual observations $v_t$ at each timestep $t$, the agent must predict an action $a_t$ and (if applicable) a binary mask $m_t$ over $v_t$ which highlights an object to interact with. After an episode length of $T$ timesteps, an inferred sequence of predicted actions $a_1, a_2, \cdots, a_T$ and masks $m_1, m_2, \cdots, m_T$ is considered successful if and only if the final state $S_T$ is equal to $S^*$. 

\subsubsection{Evaluation}
ALFRED employs three primary modes of evaluation, which we call \textit{action-by-action}, \textit{subgoal-by-subgoal}, and \textit{goal-by-goal}. Action-by-action evaluation is used while training the model to select the best model instance during hyperparameter search. It aims to score only the action sequence predicted by the agent by comparing it to the ground truth action sequence. To facilitate this evaluation, at each timestep, we simply supply the model with the ground truth action sequence up until the timestep, and compare the model's output action with the ground truth action. This can give us some ideas on how well the model can choose the next primitive action based on language instructions and the ground truth episode history. 

To evaluate the composition of both an agent's action and object interaction mask prediction in achieving particular subgoals, we use subgoal-by-subgoal evaluation.
Here, we run the agent and underlying model through the ground truth episode up until a particular subgoal, then allow the agent to attempt to complete the subgoal. The episode will continue until the agent successfully completes the subgoal, predicts stop, or reaches a time limit. The agent will be considered successful if all subgoal conditions are satisfied, e.g., if the instructions direct the agent to pick up an object, that the object is now held by the agent.

To measure how well the agent's sequential predictions for all subgoals achieve the overall goal in each ALFRED task, we use goal-by-goal evaluation, the primary mode used for ranking on the ALFRED leaderboard.\footnote{\url{https://leaderboard.allenai.org/alfred/submissions/public}} It places the agent into the virtual environment under the initial conditions, and then requires the agent to execute all steps to achieve all subgoals toward a goal. The episode ends when the agent predicts a stop action, reaches some limit of timesteps, or encounters some maximum number of API errors when interacting with the AI2 Thor environment. The agent is then evaluated based on whether the goal state is true in the environment. An agent is considered successful if all goal conditions are satisfied, but we can also score the agent based on the proportion of goal conditions it achieved. For example, if a goal is given as ``Rinse off a mug and place it in the coffee maker,''\footnote{Example from \citet{shridharALFREDBenchmarkInterpreting2020}.} the goal conditions may be that there exists a mug in the coffee maker, and that mug is clean.

\subsection{Baseline Approach}

\begin{figure}
    \centering
    \includegraphics[width=0.45\textwidth]{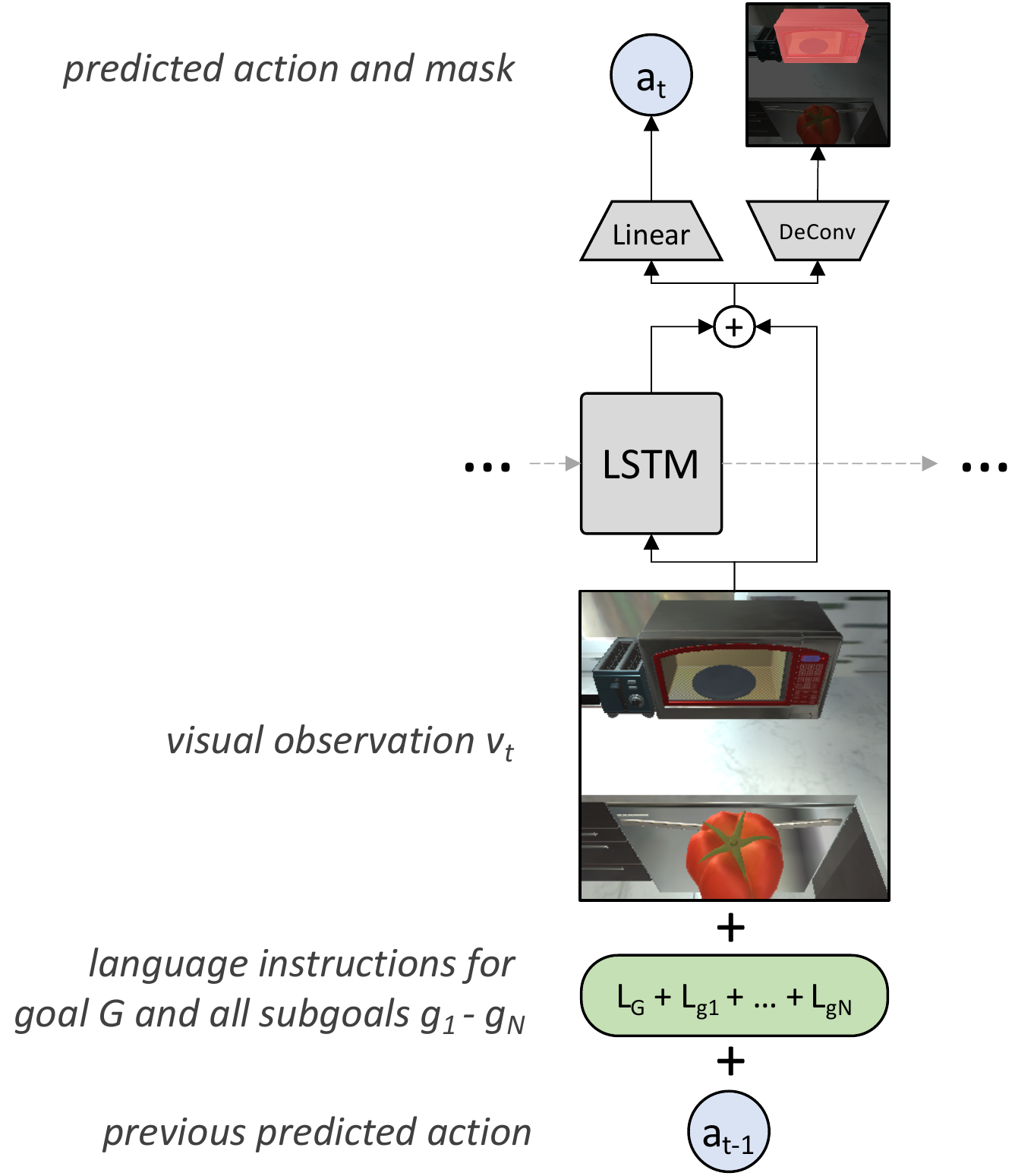}
    \caption{At each timestep, the \textsc{Seq2Seq} model takes several encoded inputs: the visual observation $v_t$, all language instructions $L_G + L_{1} + \cdots + L_{N}$ (reweighted by an attention mechanism at every timestep), and the previous predicted action $a_{t-1}$. These encoded inputs are passed into an LSTM, which generates a history-dependent representation. The original inputs and this representation can be passed to a linear layer to predict a primitive action to take and (if a manipulation action is predicted) to a deconvolutional layer to predict a binary mask over the target object.}
    \label{fig:baseline}
\end{figure}

The baseline \textsc{Seq2Seq} approach is shown in Figure~\ref{fig:baseline}. First, we generate encoded representations of all language instructions for the example, the current visual observation image, and the previously predicted action. 
Language is encoded in a bidirectional LSTM~\cite{Hochreiter:1997:LSM:1246443.1246450}, images are encoded in a frozen pre-trained ResNet~\cite{he2016deep} model, and an embedding is learned for actions.
At each timestep, these inputs are passed to a decoder LSTM. The LSTM's hidden state along with these task inputs are used to predict a primitive action to take (through a linear layer), and a binary mask to be used if a manipulation action is predicted (through a deconvolutional layer).
More detailed information on the implementation of the baseline model is provided by \citet{shridharALFREDBenchmarkInterpreting2020}.

\subsubsection{Key Limitations}
We identify several key opportunities for improvement of the baseline approach. First, the \textbf{required sequence of actions is typically quite long} before the goal state is achieved, requiring the completion of several subgoals, which each consist of several primitive actions. Such long-distance dependency is very challenging for current neural networks.

Second, the agent's \textbf{performance in navigation is a bottleneck to the overall performance}. Nearly every other subgoal in ALFRED task instances requires the agent to navigate to a new location; if the agent failed to arrive at the correct location, it will not find the objects it must manipulate afterward. Using the pre-trained weights of the baseline agent,\footnote{\url{https://github.com/askforalfred/alfred}} we found that navigation subgoals achieved a subgoal-by-subgoal success rate of 61.9\%, relatively low compared to many manipulation subgoal types, such as to cool or toggle an object, which both achieved over 90\%. As navigation occurs before every manipulation subgoal and thus is a prerequisite to all other subgoal types, if we can improve the navigation performance, then we expect to improve the overall performance of the model.

We suspect that the navigation performance is so low for two reasons. First, because the baseline model is trained by imitation learning, there are no opportunities to explore during training. During navigation, a task that relies heavily on exploration, this becomes a challenge. The agent is trained only on expert demonstrations where the agent knows exactly where to go, even if the initial location is far away or out of view of the destination. However, if the agent encounters these conditions at inference time, it is unclear whether it could have learned any exploratory behavior to be able to find the destination.

Another possible factor is that the agent is not given any explicit training toward language grounding during navigation subgoals. While the agent is trained to generate binary masks over the target objects of manipulation actions during manipulation subgoals, there is no such supervision available during navigation. Human navigation relies on first identifying landmarks, then finding routes between them \cite{siegel1975development}. Navigation instructions in ALFRED often point out landmarks along the way, for example, ``Turn around and walk towards the \textit{bed}, then hang a left and walk up to the \textit{wall}, turn left again and walk over to the right side of the wooden \textit{desk}.'' Therefore, it is advantageous for the agent to identify them, especially if the landmark refers to the destination.

\section{Proposed Improvements}
Considering the key limitations of the baseline model, we propose three main improvements. First, in order to lighten the load on the LSTM used to decode the sequence of actions, we train the model to execute only one subgoal at a time rather than the entire sequence of subgoals. Second, to enable better visual understanding and language grounding, we augmented the task inputs with additional visual observations for panoramic view angles at each timestep, and applied an object detection module to all of the agent's visual observations. Lastly, to guide navigation, we use these augmented inputs to predict an additional input to the model at every timestep: the angle between the agent's view and the goal location. This input is predicted using a new \textit{localizer} module before being passed into the LSTM.

\begin{figure*}
    \centering
    \includegraphics[width=0.95\textwidth]{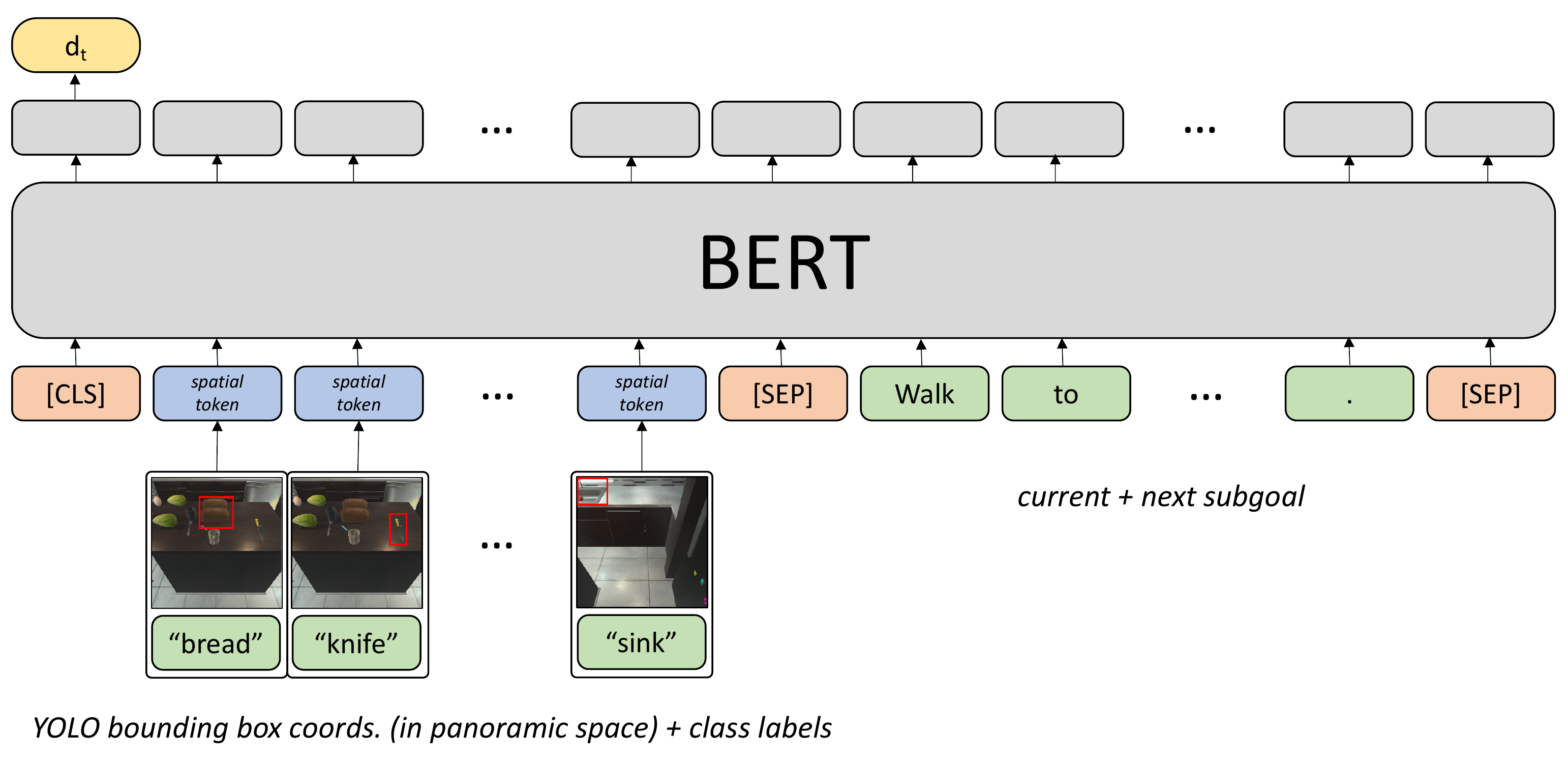}
    \caption{For a navigation subgoal $g_k$, the multimodal \textsc{BERT}-based localizer takes text and spatial tokens as input. The text input consists of language instructions $L_k$ and $L_{k+1}$ for the current and next subgoal, while the spatial input is comprised of the coordinates (projected into panoramic space) and \textsc{BERT}-embedded object class labels of panoramic bounding boxes at timestep $t$. The pre-trained \textsc{BERT} model generates a spatial representation of the model's percepts and language instructions, and we input this representation for the special  {\lbrack CLS\rbrack} token into a linear layer to predict $d_t$, the angle from the agent's orientation to the goal location.}
    \label{fig:localizer}
\end{figure*}

\subsection{Granular Training}
The LSTM-based \textsc{Seq2Seq} baseline for ALFRED is trained to predict a sequence of primitive actions from a long sequence of text (consisting of language descriptions of the overall goal and all subgoals), visual observations of the agent, and previously executed actions. However, most subgoals are disentangled from one another with minimal cross-referencing. As such, we reduced the strain on the LSTM-based model by breaking the execution down into one subgoal at a time: for each instance of inference, the model will receive the language instructions for one subgoal, and output the actions and interaction masks required to complete the subgoal.
As the language instructions are delimited by subgoal at both training and test time, we can simply iterate through them to facilitate this.

\subsection{Augmented Navigation}\label{sec:lookaround}
We augment the input data for ALFRED in two ways. First, for navigation subgoals, we add panoramic image observations to the visual input at each timestep. Second, we train an object detection model to identify objects in all visual input images.

\subsubsection{Panoramic Visual Observations}
As the baseline agent is trained by imitation learning, it does not learn any exploratory behavior. During navigation, this becomes problematic. The agent can only see directly in front of itself, so if the referred landmarks or destination are not in view, it may cause difficulties linking language to visual perception.

To mitigate this, for all navigation subgoals in the training data, we use AI2 Thor~\cite{ai2thor} to generate the agent's visual observations at eight rotation angles for each timestep. 
These observations can be used to approximate a panoramic view for the agent at each timestep during navigation, which has proven beneficial in the related vision-and-language navigation (VLN) task \cite{mattersim,friedSpeakerFollowerModelsVisionandLanguage2018}. 
In training, this allows the agent to effectively ``look around'' before taking a step. During inference, the additional observations can be generated by forcing the agent to take eight consecutive rotation actions at each timestep.

\subsubsection{Object Detection Model Training}
While the baseline agent is trained to explicitly identify objects from language instructions during object manipulation subgoals, there is no such training during navigation subgoals. This inhibits the agent's ability to ground navigation instructions to its surroundings.

To enable better language grounding, especially during navigation, we introduced YOLO V4~\cite{bochkovskiy2020yolov4}, an object detection model, to the pipeline. The model was trained on robot view images, with bounding boxes supplied by the AI2 Thor. Once trained, this object detection model enables the agent to identify 203 categories of objects in view during inference.

\begin{figure*}
    \centering
    \includegraphics[width=0.85\textwidth]{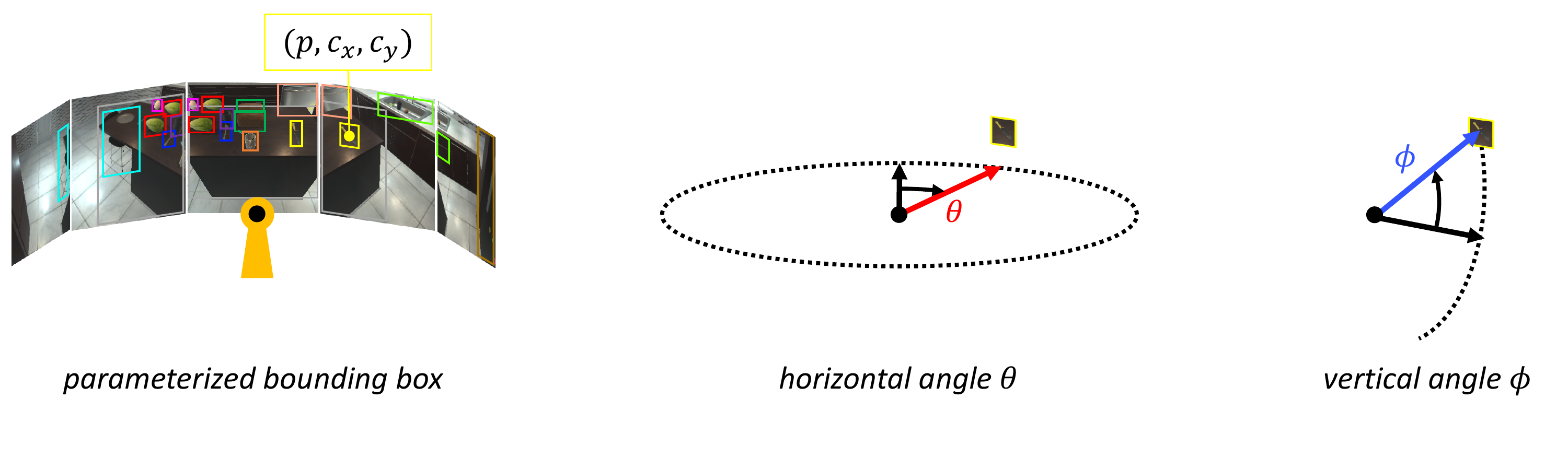}
    \caption{Spatial representation of the bounding box for a \textit{knife}. At each timestep, the agent collects visual inputs for eight panoramic view angles, and applies the trained object detection model to them. Each resulting bounding box can be parameterized by the ordinal view angle index $p$ of its image and the centroid coordinates $(c_x. c_y)$ of the bounding box within the image. We project the parametric representation for each bounding box into panoramic space by using it to calculate the horizontal angle $\theta$ and vertical angle $\phi$. }
    \label{fig:lookaround_params}
\end{figure*}

\subsection{Localizer Module}

Lastly, we combine these new inputs to introduce a new module to the pipeline: the \textit{localizer}. At each timestep during navigation subgoals, this module predicts a spatial vector $d_t$ representing the angle between the agent's orientation and direction to the goal location. 
$d_t$ is then appended to the \textsc{Seq2Seq} model's inputs at each timestep to guide navigation toward the goal location.
During non-navigation subgoals, a zero vector is passed instead.

\subsubsection{Inputs}
The agent must predict the angle toward the goal location using task inputs that are available at inference time. These include only the language instructions and any visual observations resulting from taken actions.

In order to characterize the goal location of a navigation subgoal and subsequently predict its direction, it is critical to consider the target objects of both the navigation subgoal and the following object manipulation subgoal. Consider two consecutive instructions: ``Walk to the counter'' and ``Pick up the knife''. The instruction for the navigation subgoal refers to a large object, \textit{the counter}, as the destination. If we only look at this instruction, anywhere around the counter could be a correct location. However, from the next instruction, we know that the agent should stop at a location that it can reach for \textit{the knife}.

As such, given the language instructions $L_k$ and $L_{k+1}$ for consecutive navigation and object manipulation subgoals $g_k$ and $g_{k+1}$, along with the panoramic visual observations and bounding boxes for all visible objects as described earlier, we aim to predict the direction from the agent's view to the goal location. As ALFRED provides the ground truth sequence of agent movements and the final goal location, it is straightforward to calculate the ground truth for $d_t$.
To successfully perform this task, the localizer must learn how the position and size of bounding boxes correspond to the direction and distance of objects around the agent, and how these properties may be affected by object type\footnote{For example, at the same distance, the expected sizes of a knife and of a dining table in the agent's field of view are different.} and context.\footnote{For example, if two objects of the same type are visible at once, only the context given by the language instructions can disambiguate the correct destination, e.g., ``The knife on the left.''}


\begin{table*}
    \centering
    \begin{tabular}{P{2cm}|P{1.8cm}|P{1.8cm}|P{1.8cm}|P{1.8cm}|P{1.8cm}|P{1.8cm}}\toprule
             \textbf{Model} & 
             \multicolumn{2}{|c|}{\textbf{\thead{Action-by-Action \\${F_1}$ (\%)}}} & 
             \multicolumn{2}{|c|}{\textbf{\thead{Navigation Subgoal\\Success Rate (\%)}}} & 
             \multicolumn{2}{|c}{\textbf{\thead{Goal Condition\\Success Rate (\%)}}}  \\\midrule
              & \textit{Val. Seen} & \textit{Val. Unseen}
              & \textit{Val. Seen} & \textit{Val. Unseen}
              & \textit{Val. Seen} & \textit{Val. Unseen} \\\midrule
             baseline 
             & 84.5 & 75.6 & \textbf{31.0} & 27.5 & \textbf{1.6} & 0.0 \\\midrule
             step-by-step 
             & 91.6 & 85.3 & 30.0 & 26.5 & 1.3 & 0.0 \\\midrule
             oracle 
             & \underline{93.9} & 86.9 & \underline{67.8} & \underline{35.4} & \underline{2.8} & 0.0 \\\midrule
             localizer 
             & \textbf{93.8} & \underline{\textbf{88.7}} & 25.4 & \textbf{28.8} & 1.4 & 0.0 \\\bottomrule
    \end{tabular}
    \caption{Results for the compared models. Best non-oracle result in bold, and best overall result underlined.}
    \label{tbl:results}
\end{table*}

\subsubsection{Implementation}
The full architecture of the localizer is shown in Figure~\ref{fig:localizer}. It is implemented using a multimodal variant of \textsc{BERT}~\cite{devlin-etal-2019-bert} which encodes the panoramic bounding boxes available at each timestep along with the current and next subgoal language instructions to produce a spatial representation of the agent's observations at each timestep. We perform a regression on the model output to estimate the current angle to the goal location. This model incorporates techniques proposed by \citet{li2020oscar}, who input image region features into \textsc{BERT} alongside language features for impressive results on various multimodal tasks, and \citet{miyazawa2020lambert}, who encode spatial information as input to a similar multimodal \textsc{BERT} model.

\paragraph{Bounding box coordinates in panoramic space.}
As shown in Figure~\ref{fig:lookaround_params}, the augmented navigation inputs give us labeled bounding boxes for images from eight panoramic view angles. In order to uniquely represent the position of each bounding box with respect to the agent, we project it into three-dimensional, panoramic space through two polar coordinates. In this space, we represent a bounding box by its horizontal angle $\theta$ and vertical angle $\phi$. $\theta$ will be calculated relative to the agent's body orientation, while $\phi$ will be relative to the center of the agent's vertical range of head motion.

To calculate these angles, we first parameterize each bounding box by three values: the ordinal index $p$ of the panoramic view angle for the image the bounding box belongs in, and the centroid coordinates $(c_x, c_y)$ of the bounding box within its image normalized by the image width and height. Note that two consecutive panoramic views have an angle of 45 degrees between them. Given these parameters and the horizontal field of view $F_x$, we can calculate the horizontal angle $\theta$ of a particular bounding box (in degrees) by

\begin{equation}
\theta = \arctan{\left[2(c_x-0.5)\tan{\frac{F_x}{2}}\right]} + 45p
\end{equation}

Additionally given the vertical field of view $F_y$ and the agent's current vertical heading angle $\delta$, we can calculate the vertical $\phi$ of a particular bounding box (in degrees) by

\begin{equation}
\phi = \arctan{\left[ 2(0.5-c_y)\tan{\frac{F_y}{2}}\right]} + \delta
\end{equation}

\paragraph{Input and output details.}
Once we have calculated $\theta$ and $\phi$, we encode the spatial representation as a 5-dimensional vector of $\sin{\theta}$, $\cos{\theta}$, $\sin{\phi}$, and the width $w$ and height $h$ of the bounding box, respectively. By incorporating sines and cosines, we account for the circular nature of angles and restrict the range of the values. This encoding is repeated to be comparable size to a \textsc{BERT} embedding, then added to the \textsc{BERT} embedding for the bounding box class label, e.g., \textit{knife}, to create a spatial token ready for input to \textsc{BERT}. As shown in Figure~\ref{fig:localizer}, the full input to the localizer is a concatenation of the special {\lbrack CLS\rbrack} token, the spatial tokens, the {\lbrack SEP\rbrack} token, the \textsc{BERT} embedding for the current and next subgoal language instructions $L_k$ and $L_{k+1}$, and another {\lbrack SEP\rbrack} token.

After being processed by \textsc{BERT} into a spatial representation for the agent's inputs, we use a linear layer to predict $d_t$, the sine and cosine of the angle toward the goal location. Specifically, we use the generated representation for the special {\lbrack CLS\rbrack} token. Again, sine and cosine are used to account for the circular nature of angles.

\section{Empirical Results}

Due to prohibitive training time, we conduct experiments on examples from one task type \textit{stack and place}, where each instance consists of both navigation actions and manipulation actions on multiple objects. We compare the performance of the following approaches:
\begin{itemize}
    \item \textsc{Seq2Seq} baseline (\textit{baseline})
    \item \textsc{Seq2Seq} baseline + granular training (\textit{step-by-step})
    \item \textsc{Seq2Seq} baseline + granular training + oracle spatial tracking (\textit{oracle})
    \item \textsc{Seq2Seq} baseline + granular training + localizer spatial tracking (\textit{localizer})
\end{itemize}

For the remainder of this section, we introduce the oracle approach, then present the results.

\paragraph{Oracle spatial tracking.}
To gauge the effectiveness of this spatial tracking approach and compare our model to a perfect model, we introduce an oracle spatial tracking model. At each timestep during navigation subgoals, rather than predict the angle toward the goal location $d_t$, we feed the ground truth value of this into the model directly. This gives us an upper bound on the performance of this spatial tracking approach.

\paragraph{Metrics.}
At inference time, we evaluate the model with metrics in the three increasingly strict modes of evaluation for our models derived from \citet{shridharALFREDBenchmarkInterpreting2020}: action-by-action, subgoal-by-subgoal, and goal-by-goal. To judge the agent's prediction of primitive actions, we use the \textit{action-by-action F-measure} of the predicted sequences of primitive actions for entire goal trajectories compared to the ground truth.
To judge navigation performance, we also calculate the \textit{navigation subgoal success rate}, i.e., the percentage of navigation subgoals the agent successfully completes.
As we trained our models on a relatively small subset of the data, no models achieve a viable goal completion success rate in this evaluation. Consequently, to judge the ability of the agent to achieve the overall goal of ALFRED tasks, we only report the \textit{goal condition success rate}, the percentage of all goal conditions achieved by the model.

All metrics are calculated for ALFRED's two validation sets: validation \textit{seen} and \textit{unseen}. The seen dataset uses virtual rooms from AI2 Thor which were seen in training, while the unseen uses rooms that were not seen in training.

\paragraph{Results interpretation.}
The evaluation results for the compared models are listed in Table~\ref{tbl:results}. We see that introducing granular training in the step-by-step model gives us a significant improvement in the prediction of individual actions, but achieves comparable results for other metrics. While the oracle spatial tracking approach offers drastic improvements in navigation subgoal and goal performance, the non-oracle \textsc{BERT}-based localizer approach does not come close to this upper bound. Nonetheless, it provides an overall net improvement in action-by-action F-measure of 9.3\% and 13.1\% over the baseline for seen and unseen rooms, respectively, and provides a slight improvement in navigation subgoal success rate over the baseline. This may indicate that in the task of predicting the spatial direction of targets, the transformer-based model still has much room for improvement.


\section{Conclusion and Future Work}
In this work, we investigated several methods to improve the ALFRED baseline \textsc{Seq2Seq} model, including subgoal-by-subgoal granular training, augmenting navigation inputs with panoramic visual observation images and a full coverage of object segmentation masks, and combining these new inputs to propose a novel \textsc{BERT}-based spatial tracking module. 

While granular training considerably improves precision of predicted actions, we show through an oracle approach that if the agent is given the angle toward the goal location as input at every timestep, we can achieve astounding performance improvements in navigation and goal completion. This suggests that improving navigation performance can indeed be a key to solving this problem. Our fair, learning-based spatial tracking approach makes some progress toward this grand goal of closing the gap to oracle performance, but we expect there is much more work to be done.
Beyond this, future work may include extending \textsc{BERT} or similar transformer-based models to handle the entire task end-to-end, including action and object interaction mask prediction.

\bibliography{alfred}

\begin{thebibliography}{25}
\providecommand{\natexlab}[1]{#1}
\providecommand{\url}[1]{\texttt{#1}}
\providecommand{\urlprefix}{URL }
\expandafter\ifx\csname urlstyle\endcsname\relax
  \providecommand{\doi}[1]{doi:\discretionary{}{}{}#1}\else
  \providecommand{\doi}{doi:\discretionary{}{}{}\begingroup
  \urlstyle{rm}\Url}\fi

\bibitem[{Anderson et~al.(2018)Anderson, Wu, Teney, Bruce, Johnson,
  S{\"u}nderhauf, Reid, Gould, and {van den Hengel}}]{mattersim}
Anderson, P.; Wu, Q.; Teney, D.; Bruce, J.; Johnson, M.; S{\"u}nderhauf, N.;
  Reid, I.; Gould, S.; and {van den Hengel}, A. 2018.
\newblock Vision-and-{{Language Navigation}}: {{Interpreting}}
  Visually-Grounded Navigation Instructions in Real Environments.
\newblock In \emph{Proceedings of the {{IEEE}} Conference on Computer Vision
  and Pattern Recognition ({{CVPR}})}.

\bibitem[{Bochkovskiy, Wang, and Liao(2020)}]{bochkovskiy2020yolov4}
Bochkovskiy, A.; Wang, C.-Y.; and Liao, H.-Y.~M. 2020.
\newblock YOLOv4: Optimal Speed and Accuracy of Object Detection.
\newblock \emph{arXiv preprint arXiv:2004.10934} .

\bibitem[{Chang et~al.(2017)Chang, Dai, Funkhouser, Halber, Niessner, Savva,
  Song, Zeng, and Zhang}]{changMatterport3DLearningRGBD2017}
Chang, A.; Dai, A.; Funkhouser, T.; Halber, M.; Niessner, M.; Savva, M.; Song,
  S.; Zeng, A.; and Zhang, Y. 2017.
\newblock {{Matterport3D}}: {{Learning}} from {{RGB}}-{{D Data}} in {{Indoor
  Environments}}.
\newblock \emph{International Conference on 3D Vision (3DV)} .

\bibitem[{Das et~al.(2018)Das, Datta, Gkioxari, Lee, Parikh, and
  Batra}]{embodiedqa}
Das, A.; Datta, S.; Gkioxari, G.; Lee, S.; Parikh, D.; and Batra, D. 2018.
\newblock {E}mbodied {Q}uestion {A}nswering.
\newblock In \emph{Proceedings of the IEEE Conference on Computer Vision and
  Pattern Recognition (CVPR)}.

\bibitem[{Devlin et~al.(2019)Devlin, Chang, Lee, and
  Toutanova}]{devlin-etal-2019-bert}
Devlin, J.; Chang, M.-W.; Lee, K.; and Toutanova, K. 2019.
\newblock {BERT}: Pre-training of Deep Bidirectional Transformers for Language
  Understanding.
\newblock In \emph{Proceedings of the 2019 Conference of the North {A}merican
  Chapter of the Association for Computational Linguistics: Human Language
  Technologies, Volume 1 (Long and Short Papers)}, 4171--4186. Minneapolis,
  Minnesota: Association for Computational Linguistics.
\newblock \doi{10.18653/v1/N19-1423}.
\newblock \urlprefix\url{https://www.aclweb.org/anthology/N19-1423}.

\bibitem[{Fried et~al.(2018)Fried, Hu, Cirik, Rohrbach, Andreas, Morency,
  {Berg-Kirkpatrick}, Saenko, Klein, and
  Darrell}]{friedSpeakerFollowerModelsVisionandLanguage2018}
Fried, D.; Hu, R.; Cirik, V.; Rohrbach, A.; Andreas, J.; Morency, L.-P.;
  {Berg-Kirkpatrick}, T.; Saenko, K.; Klein, D.; and Darrell, T. 2018.
\newblock Speaker-{{Follower Models}} for {{Vision}}-and-{{Language
  Navigation}}.
\newblock \emph{arXiv:1806.02724 [cs]} .

\bibitem[{He et~al.(2016)He, Zhang, Ren, and Sun}]{he2016deep}
He, K.; Zhang, X.; Ren, S.; and Sun, J. 2016.
\newblock Deep residual learning for image recognition.
\newblock In \emph{Proceedings of the IEEE conference on computer vision and
  pattern recognition}, 770--778.

\bibitem[{Hochreiter and
  Schmidhuber(1997)}]{Hochreiter:1997:LSM:1246443.1246450}
Hochreiter, S.; and Schmidhuber, J. 1997.
\newblock Long {{Short}}-{{Term Memory}}.
\newblock \emph{Neural Computation} 9(8): 1735--1780.
\newblock ISSN 0899-7667.
\newblock \doi{10.1162/neco.1997.9.8.1735}.

\bibitem[{Hu et~al.(2019)Hu, Fried, Rohrbach, Klein, Darrell, and
  Saenko}]{hu-etal-2019-looking}
Hu, R.; Fried, D.; Rohrbach, A.; Klein, D.; Darrell, T.; and Saenko, K. 2019.
\newblock Are You Looking? Grounding to Multiple Modalities in
  Vision-and-Language Navigation.
\newblock In \emph{Proceedings of the 57th Annual Meeting of the Association
  for Computational Linguistics}, 6551--6557. Florence, Italy: Association for
  Computational Linguistics.
\newblock \doi{10.18653/v1/P19-1655}.
\newblock \urlprefix\url{https://www.aclweb.org/anthology/P19-1655}.

\bibitem[{Jain et~al.(2019)Jain, Magalhaes, Ku, Vaswani, Ie, and
  Baldridge}]{jain-etal-2019-stay}
Jain, V.; Magalhaes, G.; Ku, A.; Vaswani, A.; Ie, E.; and Baldridge, J. 2019.
\newblock Stay on the Path: Instruction Fidelity in Vision-and-Language
  Navigation.
\newblock In \emph{Proceedings of the 57th Annual Meeting of the Association
  for Computational Linguistics}, 1862--1872. Florence, Italy: Association for
  Computational Linguistics.
\newblock \doi{10.18653/v1/P19-1181}.
\newblock \urlprefix\url{https://www.aclweb.org/anthology/P19-1181}.

\bibitem[{Kolve et~al.(2017)Kolve, Mottaghi, Han, VanderBilt, Weihs, Herrasti,
  Gordon, Zhu, Gupta, and Farhadi}]{ai2thor}
Kolve, E.; Mottaghi, R.; Han, W.; VanderBilt, E.; Weihs, L.; Herrasti, A.;
  Gordon, D.; Zhu, Y.; Gupta, A.; and Farhadi, A. 2017.
\newblock {{AI2}}-{{THOR}}: {{An}} Interactive {{3D}} Environment for Visual
  {{AI}}.
\newblock \emph{arXiv} .

\bibitem[{Li et~al.(2020)Li, Yin, Li, Hu, Zhang, Zhang, Wang, Hu, Dong, Wei,
  Choi, and Gao}]{li2020oscar}
Li, X.; Yin, X.; Li, C.; Hu, X.; Zhang, P.; Zhang, L.; Wang, L.; Hu, H.; Dong,
  L.; Wei, F.; Choi, Y.; and Gao, J. 2020.
\newblock Oscar: Object-Semantics Aligned Pre-training for Vision-Language
  Tasks.
\newblock \emph{arXiv preprint arXiv:2004.06165} .

\bibitem[{Ma et~al.(2019{\natexlab{a}})Ma, Lu, Wu, AlRegib, Kira, Socher, and
  Xiong}]{ma2019selfmonitoring}
Ma, C.-Y.; Lu, J.; Wu, Z.; AlRegib, G.; Kira, Z.; Socher, R.; and Xiong, C.
  2019{\natexlab{a}}.
\newblock Self-Monitoring Navigation Agent via Auxiliary Progress Estimation.
\newblock In \emph{Proceedings of the International Conference on Learning
  Representations (ICLR)}.
\newblock \urlprefix\url{https://arxiv.org/abs/1901.03035}.

\bibitem[{Ma et~al.(2019{\natexlab{b}})Ma, Wu, AlRegib, Xiong, and
  Kira}]{ma2019theregretful}
Ma, C.-Y.; Wu, Z.; AlRegib, G.; Xiong, C.; and Kira, Z. 2019{\natexlab{b}}.
\newblock The Regretful Agent: Heuristic-Aided Navigation through Progress
  Estimation.
\newblock In \emph{Proceedings of the IEEE Conference on Computer Vision and
  Pattern Recognition (CVPR)}.
\newblock \urlprefix\url{https://arxiv.org/abs/1903.01602}.

\bibitem[{Majumdar et~al.(2020)Majumdar, Shrivastava, Lee, Anderson, Parikh,
  and Batra}]{majumdarImprovingVisionandLanguageNavigation2020}
Majumdar, A.; Shrivastava, A.; Lee, S.; Anderson, P.; Parikh, D.; and Batra, D.
  2020.
\newblock Improving {{Vision}}-and-{{Language Navigation}} with
  {{Image}}-{{Text Pairs}} from the {{Web}}.
\newblock \emph{arXiv:2004.14973 [cs]} .

\bibitem[{Miyazawa et~al.(2020)Miyazawa, Aoki, Horii, and
  Nagai}]{miyazawa2020lambert}
Miyazawa, K.; Aoki, T.; Horii, T.; and Nagai, T. 2020.
\newblock lamBERT: Language and Action Learning Using Multimodal BERT.
\newblock \emph{arXiv preprint arXiv:2004.07093} .

\bibitem[{Qi et~al.(2020)Qi, Wu, Anderson, Wang, Wang, Shen, and van~den
  Hengel}]{reverie}
Qi, Y.; Wu, Q.; Anderson, P.; Wang, X.; Wang, W.~Y.; Shen, C.; and van~den
  Hengel, A. 2020.
\newblock REVERIE: Remote Embodied Visual Referring Expression in Real Indoor
  Environments.
\newblock In \emph{Proceedings of the IEEE Conference on Computer Vision and
  Pattern Recognition (CVPR)}.

\bibitem[{Savva et~al.(2019)Savva, Kadian, Maksymets, Zhao, Wijmans, Jain,
  Straub, Liu, Koltun, Malik, Parikh, and Batra}]{habitat19iccv}
Savva, M.; Kadian, A.; Maksymets, O.; Zhao, Y.; Wijmans, E.; Jain, B.; Straub,
  J.; Liu, J.; Koltun, V.; Malik, J.; Parikh, D.; and Batra, D. 2019.
\newblock Habitat: {A} {P}latform for {E}mbodied {AI} {R}esearch.
\newblock In \emph{Proceedings of the IEEE/CVF International Conference on
  Computer Vision (ICCV)}.

\bibitem[{Shridhar et~al.(2020)Shridhar, Thomason, Gordon, Bisk, Han, Mottaghi,
  Zettlemoyer, and Fox}]{shridharALFREDBenchmarkInterpreting2020}
Shridhar, M.; Thomason, J.; Gordon, D.; Bisk, Y.; Han, W.; Mottaghi, R.;
  Zettlemoyer, L.; and Fox, D. 2020.
\newblock {{ALFRED}}: {{A Benchmark}} for {{Interpreting Grounded
  Instructions}} for {{Everyday Tasks}}.
\newblock In \emph{Computer {{Vision}} and {{Pattern Recognition}} ({{CVPR}})}.

\bibitem[{Siegel and White(1975)}]{siegel1975development}
Siegel, A.~W.; and White, S.~H. 1975.
\newblock The Development of Spatial Representations of Large-Scale
  Environments.
\newblock In \emph{Advances in Child Development and Behavior}, volume~10,
  9--55. {Elsevier}.

\bibitem[{Thomason et~al.(2019)Thomason, Murray, Cakmak, and
  Zettlemoyer}]{thomason:arxiv19}
Thomason, J.; Murray, M.; Cakmak, M.; and Zettlemoyer, L. 2019.
\newblock Vision-and-Dialog Navigation.
\newblock In \emph{2019 Conference on Robot Learning (CoRL 2019)}.

\bibitem[{Xia et~al.(2019)Xia, Li, Chen, Shen, Mart{\'i}n-Mart{\'i}n, Hirose,
  Zamir, Fei-Fei, and Savarese}]{xiagibson2019}
Xia, F.; Li, C.; Chen, K.; Shen, W.~B.; Mart{\'i}n-Mart{\'i}n, R.; Hirose, N.;
  Zamir, A.~R.; Fei-Fei, L.; and Savarese, S. 2019.
\newblock Gibson Env V2: Embodied Simulation Environments for Interactive
  Navigation.
\newblock Technical report, Stanford University.

\bibitem[{{Xia} et~al.(2020){Xia}, {Shen}, {Li}, {Kasimbeg}, {Tchapmi},
  {Toshev}, {Martín-Martín}, and {Savarese}}]{igibson}
{Xia}, F.; {Shen}, W.~B.; {Li}, C.; {Kasimbeg}, P.; {Tchapmi}, M.~E.; {Toshev},
  A.; {Martín-Martín}, R.; and {Savarese}, S. 2020.
\newblock Interactive Gibson Benchmark: A Benchmark for Interactive Navigation
  in Cluttered Environments.
\newblock \emph{IEEE Robotics and Automation Letters} 5(2): 713--720.

\bibitem[{Yan et~al.(2018)Yan, Misra, Bennnett, Walsman, Bisk, and
  Artzi}]{yan2018chalet}
Yan, C.; Misra, D.; Bennnett, A.; Walsman, A.; Bisk, Y.; and Artzi, Y. 2018.
\newblock Chalet: Cornell house agent learning environment.
\newblock \emph{arXiv preprint arXiv:1801.07357} .

\bibitem[{Zhu et~al.(2020)Zhu, Zhu, Chang, and Liang}]{Zhu_2020_CVPR}
Zhu, F.; Zhu, Y.; Chang, X.; and Liang, X. 2020.
\newblock Vision-Language Navigation With Self-Supervised Auxiliary Reasoning
  Tasks.
\newblock In \emph{Proceedings of the IEEE/CVF Conference on Computer Vision
  and Pattern Recognition (CVPR)}.

\end{thebibliography}
\end{document}